# Multi-dimensional Data Analysis and Applications Basing on LLM Agents and Knowledge Graph Interactions

**Xi Wang**[*1,2] **Xianyao Ling**[2] **Kun Li**[2] **Gang Yin**[2] **Liang Zhang**[1,3]
**Jiang Wu**[3] **Jun Xu**[4] **Fu Zhang**[5] **Wenbo Lei**[6] **Annie Wang**[7] **Peng Gong**[8]

[1]Tsinghua University [2] Cross-strait Tsinghua Research Institute
[3]OceanBlue Construction Co. Beijing, Ltd.
[4]Hunan Jiace Evaluation Information Technology Service Co., Ltd.
[5]Huacai Technology (Beijing) Co., Ltd. [6]Beijing Yixin Technology Co., Ltd.
[7]Kalavai Corp [8]The University of Hong Kong

xi-wang19@mails.tsinghua.edu.cn xianyao.ling@ctri.org.cn likun@ctri-ai.cn yingang@ctri-ai.cn
liang-zh19@mails.tsinghua.edu.cn jiang.wu@rcytgs.com xujun@jiacetest.com
zhangfu@huacaizhaoyu.com lwb@it-hdz.com annie@kalavai.net penggong@hku.hk

**Abstract:** In the current era of big data, extracting deep insights from massive, heterogeneous, and complexly associated multi-dimensional data has become a significant challenge. Large Language Models (LLMs) perform well in natural language understanding and generation, but still suffer from "hallucination" issues when processing structured knowledge and are difficult to update in real-time. Although Knowledge Graphs (KGs) can explicitly store structured knowledge, their static nature limits dynamic interaction and analytical capabilities. Therefore, this paper proposes a multi-dimensional data analysis method based on the interactions between LLM agents and KGs, constructing a dynamic, collaborative analytical ecosystem. This method utilizes LLM agents to automatically extract product data from unstructured data, constructs and visualizes the KG in real-time, and supports users in deep exploration and analysis of graph nodes through an interactive platform. Experimental results show that this method has significant advantages in product ecosystem analysis, relationship mining, and user-driven exploratory analysis, providing new ideas and tools for multi-dimensional data analysis.

## 1. Introduction

In recent years, with the explosive growth of data volume and leaps in computing power, we are entering an era of data-driven decision-making. How to extract deep insights from massive, heterogeneous, and interrelated data has become a core challenge faced by both academia and industry. Especially for multi-dimensional data (multidimensional data), such as product information containing complex attributes, multiple relationships, and dynamic evolution characteristics, traditional analysis methods often fall short. The emergence of Large Language Models (LLMs), with their powerful natural language understanding, generation, and reasoning capabilities, has brought revolutionary changes to the field of data analysis [1][2]. From GPT-3 to its successors, LLMs have demonstrated astonishing performance in tasks such as zero-shot and few-shot learning, text generation, and code writing [3][4][5].

However, LLMs also have inherent limitations. They are prone to "hallucination" when processing highly structured knowledge requiring precise facts and complex relationships, and their internal knowledge is implicit and difficult to update in real-time [6][7]. To overcome these shortcomings, researchers have begun to explore paths combining LLMs with Knowledge Graphs (KGs) [6]. KGs, with their "entity-relationship-entity" triple form, can explicitly store and manage structured knowledge, providing factual basis and reasoning skeletons for LLMs [8][9]. This collaborative paradigm has achieved significant success in multiple fields, such as improving the accuracy of question-answering systems through KG-enhanced Retrieval-Augmented Generation (KG-RAG) [6], utilizing KGs for complex multi-hop reasoning [10], and injecting knowledge into LLMs to optimize the authenticity of their generated content [11]. Some researchers have proposed roadmaps for unifying LLMs and KGs, systematically elaborating the great potential of their integration [12][13].

---

* Corresponding author.

On this basis, a more advanced research paradigm is emerging: upgrading LLMs from passive knowledge processing tools to active Agents. LLM agents not only possess language capabilities but can also perform planning, use tools, and interact with the external environment [14][15]. Seminal work like the ReAct framework, by interleaving the generation of reasoning traces and actions, enables agents to interact with external knowledge sources like Wikipedia to complete complex tasks [16]. Subsequently, research on Multi-Agent Systems and dynamic agent networks has further expanded their collaborative and complex problem-solving capabilities, such as stimulating divergent thinking in models through debates among agents [17], or constructing dynamic agent collaboration frameworks to adapt to different query needs [18]. These advances indicate that LLM agents have the potential to become a powerful link connecting unstructured information, structured knowledge bases, and user needs. In specific domains, such as automated knowledge discovery from scientific literature, researchers have begun using dual-agent methods to construct ontologies and KGs [19].

Although existing research has made significant progress in integrating LLMs with KGs and developing LLM agents, the following shortcomings still exist:

(1) **Static Interaction Mode:** Most current research treats KGs as static, pre-built external knowledge bases, with LLMs or agents primarily performing information retrieval or verification from them [20][21]. There is a lack of a dynamic interaction loop, where agents can not only "consume" the KG but also actively "construct" and "extend" it during the interaction process, allowing the knowledge base to evolve with the analysis process.

(2) **Limitations in Analytical Dimensions:** Existing work mostly focuses on single tasks with clear objectives, such as Question Answering (QA) [10] or fact-checking. For complex scenarios requiring exploratory analysis from multiple perspectives and levels (e.g., analyzing competition, substitution, and complementary relationships among different products in a product ecosystem), a unified and flexible framework is still lacking. How to support users in truly "multi-dimensional data analysis" remains an open question.

(3) **Disjointed Human-Machine Collaboration:** Visual analysis of KGs [22][23] and reasoning analysis by LLM agents are often two separate stages. Users find it difficult to seamlessly trigger agents for deep analysis during intuitive graph exploration; conversely, the results generated by agent analysis are difficult to feedback dynamically and intuitively onto the visual interface, thus hindering smooth and efficient human-machine collaborative exploration. Existing work rarely achieves real-time, bidirectional interaction between agents and the visualized KG.

To bridge the aforementioned research gaps, this paper proposes a multi-dimensional data analysis method based on the interactions between LLM agents and KGs. This method aims to create a dynamic, collaborative analytical ecosystem for deeply parsing product information and complex relationships between different products. Our core idea is: First, utilize the powerful understanding and planning capabilities of LLM agents to automatically query and extract structured product attribute and relationship data from unstructured or semi-structured data sources (such as databases, web pages). Subsequently, based on this structured data, instantly construct a product KG and present it visually through an interactive web platform. This platform not only displays product entities (nodes) and their relationships (edges) but, more crucially, allows users to deeply interact with the KG on the visual interface. When a user is interested in a particular product node, they can trigger an integrated third-party LLM or specialized agent API with one click to conduct deeper, multi-dimensional analysis and introduction of that product, such as generating a competitive analysis report, summarizing user sentiment tendencies, or predicting market trends. This design achieves dynamic, real-time bidirectional empowerment between the LLM agent and the KG: the agent is responsible for constructing and enriching the KG, while the visualized KG serves as the "operating table" and "context" for the agent's in-depth analysis, ultimately providing the user with an unprecedented, seamless analytical experience from macro relationships to micro insights.

The main contributions of this paper can be summarized as follows:

(1) **Proposed a novel integrated analysis framework:** We integrate LLM agents, dynamic

KG construction, and interactive visualization, providing an end-to-end solution for complex multi-dimensional data analysis. This framework surpasses the traditional unidirectional modes of "KG-enhanced LLM" or "LLM-populated KG".

(2) **Designed a dynamic and real-time interaction mechanism:** We implemented a dynamic interaction loop between LLM agents and KGs. The agent is not only a knowledge acquirer but also a builder and extender of the KG. Simultaneously, users can directly "drive" the agent through the visual interface to conduct instant, in-depth exploration of any node in the graph, significantly enhancing the flexibility and timeliness of analysis.

(3) **Verified application value in the field of product analysis:** We applied this method to complex product information analysis scenarios, demonstrating its effectiveness and superiority in understanding product ecosystems, mining deep relationships between products, and empowering users to conduct independent exploratory analysis, providing a powerful new tool for business intelligence and market analysis.

## 2. Framework and Theory

### 2.1 Overall Framework

The overall framework of this method is shown in Figure 1, consisting of four core modules: the Data Preparation Module, the Knowledge Representation Module, the Visualization and Interaction Module, and the Intelligent Analysis Module.

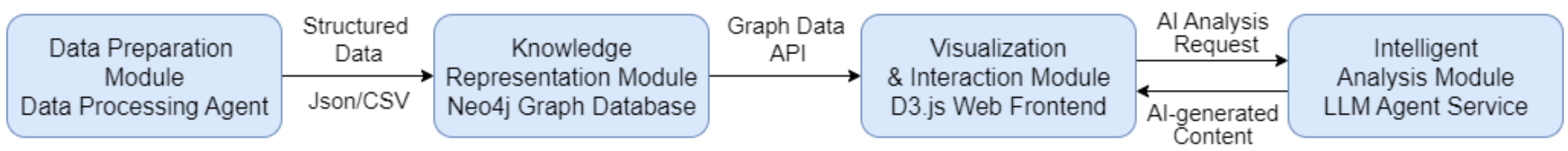

Figure 1. Overall Framework Diagram.

(1) **Data Preparation Module:** This module is primarily responsible for a data processing agent. Based on preset rules or models, this agent actively queries and extracts product information from various data sources (such as databases, web pages), parses, cleans, and converts it into a structured data format (e.g., JSON or CSV), preparing for the construction of the KG.

(2) **Knowledge Representation Module:** This module is responsible for modeling the structured data and storing it persistently. We adopt the KG as the core data representation model, using the Neo4j graph database for storage and management [24][25]. Using the Cypher query language, we load entities and relationships into the database, forming an interconnected knowledge network [24].

(3) **Visualization and Interaction Module:** This module is the main interface for user interaction with the system. We built a web frontend application based on the D3.js framework [26]. This application obtains graph data from the Neo4j database through backend APIs and renders it using layout algorithms such as the Force-Directed Graph [27][28]. Users can perform interactive operations on this interface such as search, zoom, drag, expand/collapse nodes, etc.

(4) **Intelligent Analysis Module:** This is the core innovation of this framework. This module integrates third-party LLMs and agent services via API. When a user triggers the "AI Product Summary" function on the visual interface, the system packages the key attribute information of the current node (obtained from the Knowledge Representation Module) into a carefully designed prompt, and sends it to the Intelligent Analysis Module. The LLM agent in this module, upon receiving the request, performs deep analysis, information integration, and content generation, and returns the result to the frontend for display.

### 2.2 Knowledge Graph Data Modeling

To effectively organize multi-dimensional product data, we designed a concise yet expressive KG schema. This schema defines the following core entity (node) types and relationship (edge) types [29][30]:

Node Types (Node Labels):

- Category: e.g., "Computing Server", "Central Processing Unit".
- Product: e.g., "Huawei TaiShan Server", "Intel Xeon Processor".
- Brand: e.g., "Huawei", "Intel".
- Model: e.g., "TaiShan 2280 V2", "Xeon Platinum 8375C".
- Price: e.g., "¥50000", as an entity node, facilitating future expansion of multi-currency, historical price information, etc.

Relationship Types (Relationship Types):

- Belongs_to: Connects Product nodes and Category nodes.
- Has_brand: Connects Product nodes and Brand nodes.
- Has_model: Connects Product nodes and Model nodes.
- Has_price: Connects Product nodes and Price nodes.

This graph structure model can intuitively reflect the intrinsic connections between different products. For example, through brand nodes, all products under the same brand can be easily found; through category nodes, comparisons of similar products can be conducted.

### 2.3 Agent Integration

The key implementation method in this paper lies in achieving effective linkage between the KG and LLM agents. This linkage is not a simple API call, but a context-aware, KG-driven interaction process. Its core mechanism is as follows:

(1) **Context Extraction:** When a user selects a Product node on the visual interface and requests an AI introduction, the system does not simply send the product name to the LLM. Instead, it first queries the KG for key information directly connected to this node, such as its brand, model, etc.

(2) **Prompt Engineering:** The system embeds the extracted structured information into a preset template to construct a high-quality prompt. This technique, a variant of "Chain of Thought Prompting" [31], can guide the LLM to perform more logical and in-depth thinking [32]. An example prompt might be as follows:

*"You are a seasoned IT product analyst. Please provide me with a detailed analysis report on this product based on the following structured information. The report should include its core technical features, market positioning, main application scenarios, and potential competitors.*
*Product Name: Huawei TaiShan Server*
*Brand: Huawei*
*Model: TaiShan 2280 V2"*

(3) **Dynamic Invocation and Display:** This prompt is sent to a third-party LLM agent via API. The LLM agent, based on its powerful pre-trained knowledge and combined with the precise context provided in the prompt, generates a detailed, professional analysis text. The system, upon receiving the returned result, dynamically displays it to the user in a pop-up window, completing the transformation from structured data to in-depth analytical content.

In this way, the KG provides "fact anchors" for the LLM, greatly reducing the risk of the model generating irrelevant or incorrect information (hallucination), while the LLM endows the static KG with an "intelligent brain," enabling it to provide insights beyond the data itself [33].

## 3. Method Implementation

This chapter will detail the specific implementation steps of the method in this paper, including data preparation, KG construction, visual platform development, and interactive function implementation.

### 3.1 Data Preparation

To acquire the data required for this paper, a data processing agent was constructed utilizing technologies such as LLMs, web search tools, and big data analysis. This agent can search and quickly match the most eligible several products from the vast amount of internet data based on user-input product names, specification parameters, and other key information, while also providing the corresponding brand, model, and price information for each product. The specific steps for the data processing agent to acquire product data are as follows:

(1) For various types of products (such as servers, switches, cameras, air conditioners, mobile phones, microphones, etc.), based on the user-input product name and specification parameters, first use a LLM to extract 5-7 keywords as query terms for web search.

(2) Based on the extracted keywords, combined with the web search tool API, use a method of gradually decreasing keywords to perform multiple rounds of iterative searches from the internet and return several relevant webpage data.

(3) For each relevant webpage, use an LLM to extract information from the webpage content, extracting attribute values such as product name, product type, specification parameters, brand, model, price, etc. The results are returned in a structured JSON format.

(4) Use the LLM again to judge and classify the user-input parameters and the product specification parameters extracted from the webpages (retain core parameters, remove non-core parameters), and calculate the semantic similarity between the user-input parameters and the parameters to be compared using an Embedding model [34].

(5) Filter out data missing key attribute values such as specification parameters, brand, model, price, etc., through rules, and sort them according to similarity from high to low, finally returning the top 10 results and saving the data to the database.

### 3.2 Knowledge Graph Construction

Based on the structured product data collected by the method described in Section 3.1, use the Neo4j graph database and Cypher language to construct the KG[25][35]. The specific steps are as follows:

(1) First, define the KG schema, including the node types and relationship types described in Section 2.2. Each node type has a specific set of attributes, such as product nodes containing attributes like product name, product type, and specification parameters. Different types of nodes are connected through specific relationship edges, forming a rich semantic network. This multi-level relationship network can effectively capture complex associations in the product ecosystem [24].

(2) Use the Cypher language to import the processed product data into the Neo4j graph database. For each product data item, use the MERGE statement to create or match nodes and relationships, avoiding data duplication. For example, the following Cypher code snippet shows how to create a product and its relationship with a brand:

```
MERGE (p:Product {name: 'Huawei TaiShan Server', description: 'A high-performance server based on Kunpeng processors'})
MERGE (b:Brand {name: 'Huawei'})
MERGE (p)-[:HAS_BRAND]->(b)
```

After the above steps, we constructed a small-scale product KG, containing 963 nodes (including 49 category nodes, 269 product nodes, 147 brand nodes, 265 model nodes, 233 price nodes) and 1110 relationships.

### 3.3 Visualization Platform Development

For the visualization platform, we adopted a frontend-backend separated architecture for development:

(1) **Backend Development:** Built using Node.js and the Express framework. It is responsible for providing RESTful API interfaces, receiving requests from the frontend, converting them into Cypher query statements sent to the Neo4j database, obtaining KG node and relationship data in real-time, processing the query results into the JSON format required by the frontend (usually a structure containing nodes and links arrays), and returning them.

(2) **Frontend Development:** Built using HTML, CSS, and JavaScript, with the core visualization functionality implemented by the D3.js library. D3.js can bind data to DOM elements and provides rich tools for creating dynamic, interactive data visualizations. We primarily used the forceSimulation module of D3.js to create a force-directed graph, where nodes repel each other and edges pull connected nodes closer like springs, thus forming a natural and highly readable layout [33]. The visualization effect is specially optimized, with different node types encoded using different colors and shapes to enhance visual distinction [36].

### 3.4 Interactive Function Implementation

The method in this paper implements various dynamic interactive functions: Node Query and Graph Loading, Node Information Display, Expand and Hide Nodes, Agent Product Analysis.

#### 3.4.1 Node Query and Graph Loading

A search box is located at the top of the platform. After the user enters a keyword, the frontend sends a query request to the backend API. The backend executes the following Cypher query to find nodes whose names contain the keyword and their first-order neighbors, and returns the results based on the user-set quantity limit:

*MATCH (n) WHERE n.name CONTAINS $keyword*
*MATCH (n)-[r]-(m)*
*RETURN n, r, m LIMIT $limit*

After receiving the data, the frontend clears the canvas and renders the new graph.

#### 3.4.2 Node Information Display

The platform allows viewing the attribute information of nodes and their relationships with other nodes. We bound a click event to each SVG node element in D3.js. When the user clicks on a node, the event handler reads the data bound to that node (including type, name, description, and other attributes) and displays it in the information panel in the lower left corner of the platform.

#### 3.4.3 Expand and Hide Nodes

The platform can expand or hide specific nodes and their relationship information according to user needs.

**Expand Node:** When the user right-clicks a node, a context menu pops up containing an "Expand Node" button. After clicking, the frontend sends the node's ID to the backend. The backend executes a query to obtain all neighbor nodes and relationships of this node that are not yet displayed in the frontend, and returns them to the frontend. The frontend appends these new node and relationship data to the existing dataset and restarts the D3.js force simulation, thus achieving dynamic expansion of the graph.

**Hide Node:** After clicking the "Hide Node" button, the frontend directly removes the node and its associated edges from the node and relationship data array it maintains, then updates the D3.js rendering, causing the node to smoothly disappear from view. This is a pure frontend operation with fast response speed.

#### 3.4.4 Agent Product Analysis

For each product node in the KG, the platform assists in analyzing and introducing the product information by integrating third-party LLM agents, enabling users to analyze product data of interest from multiple dimensions and realizing dynamic, real-time interaction between the agent and the KG. This is also the core interactive function of the method in this paper. Its implementation flow is as follows:

(1) **Trigger:** The user right-clicks a Product-type node and selects "AI Product Summary".

(2) **Context Extraction:** The frontend extracts the name attribute from the data object of this node and finds the name of its Brand and Model neighbor nodes.

(3) **API Call:** The frontend sends this information to the backend's /ai-introduce interface.

(4) **Backend Processing and LLM Interaction:** After receiving the request, the backend constructs a Prompt according to the prompt engineering method described in Section 2.3. Then, it sends the Prompt to a third-party LLM agent service via API call. In this implementation, we integrated the DeepSeek-V3 model API because it performs well in handling complex analysis tasks. The API call response time is typically within a few seconds, depending on the model's load [37].

(5) **Result Display:** After the backend receives the Markdown-formatted text generated by the LLM agent, it returns it to the frontend. The frontend uses an elegant Modal component to render and display the content to the user.

## 4. Experiment and Analysis

To verify the effectiveness of the method in this paper, we conducted a series of functional tests and case analyses.

The KG visualization interactive platform in this paper mainly includes the following dynamic interactive functions: Node Query and Graph Loading, Node Information Display, Expand and Hide Nodes, Agent Product Analysis. The overall effect diagram of the platform is shown in Figure 2.

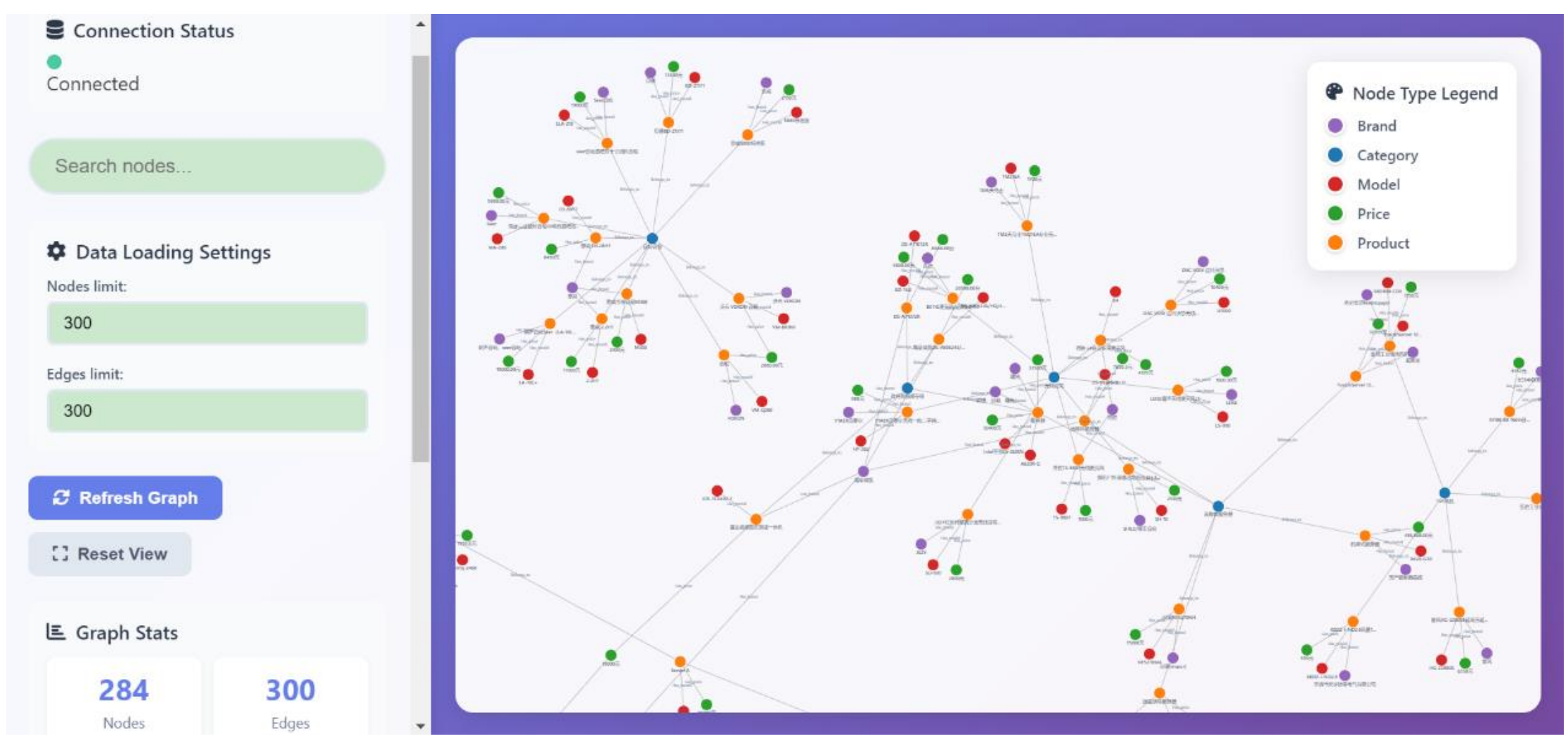

Figure 2. Overall Diagram of the KG Visualization Interactive Platform.

### 4.1 Node Query and Graph Loading

This module is located at the top of the platform, providing users with an entry point into the KG. Users can quickly locate and load subgraphs of interest by entering any node name (such as category, brand, or specific product name). Simultaneously, users can set the maximum number of nodes and relationships to be displayed during initial loading or querying via a slider or input box, which is crucial for controlling visualization complexity and ensuring frontend performance when dealing with large-scale graphs.

As shown in Figure 3, when the user enters "Computing Server" in the search box and executes the query, while setting both the node count and relationship count to 25, the system retrieves the 25 nodes and relationships most relevant to the "Computing Server" category node from the KG and displays them in the visualization area. The figure clearly shows that the "Computing Server" category includes multiple specific product nodes, and these products are respectively linked to their own brands and models.

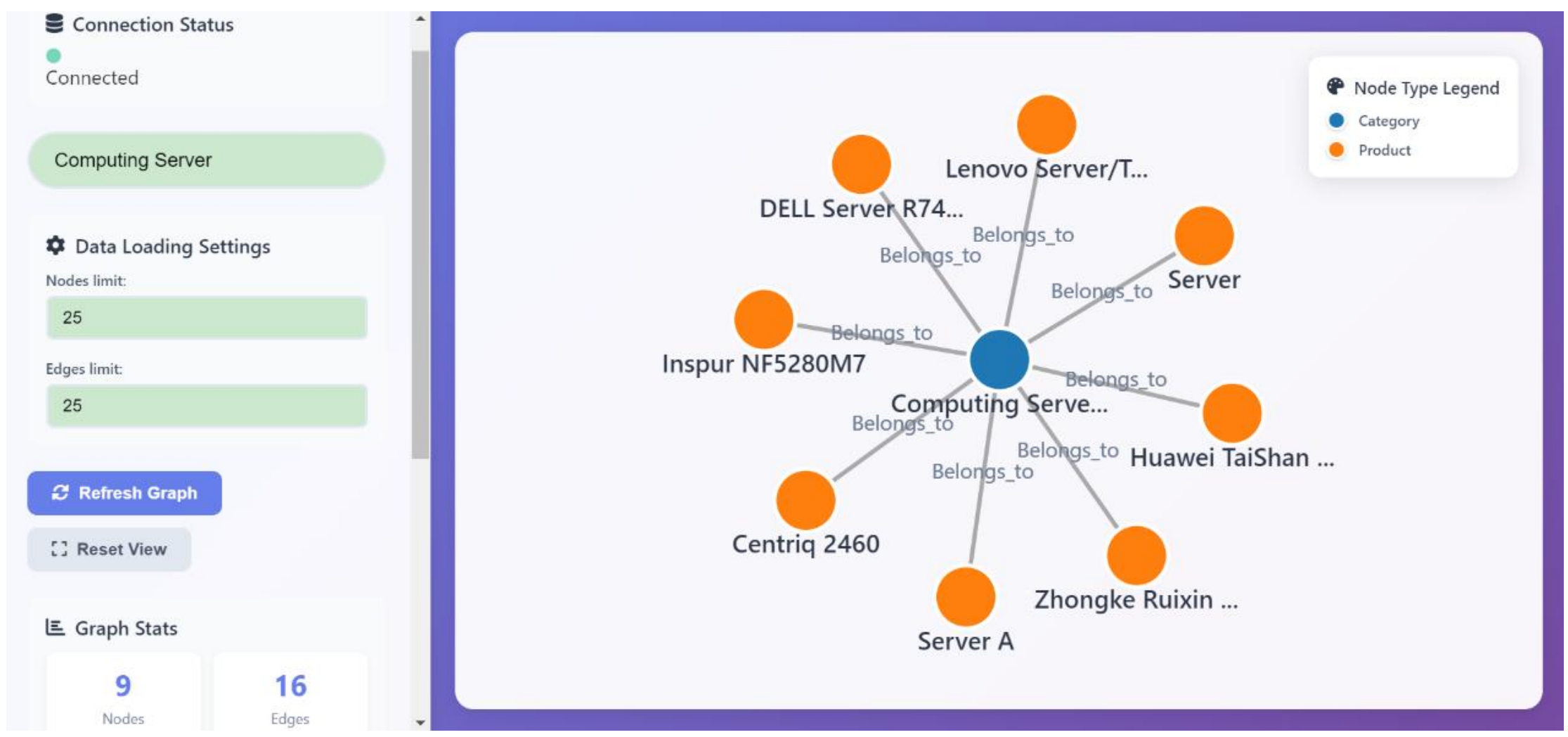

Figure 3. Example of KG Node Query and Its Relationship Graph Display.

### 4.2 Node Information Display

To allow users to understand the detailed information of each node in the graph, we designed the node information display module. When the user clicks on any node in the visualization area with the mouse, the background of that node is highlighted, and its detailed attribute information is dynamically displayed in the area at the lower left corner of the platform.

As shown in Figure 4, when the user clicks on the "Huawei TaiShan Server" product node, the information display area in the lower left corner of the platform clearly lists the node type as "Product" and displays its specific product name, belonging category, and detailed description. This instant feedback mechanism allows users to conveniently obtain the attributes of micro-entities while exploring macro relationships, conforming to the human cognitive habit of analyzing from the whole to the part.

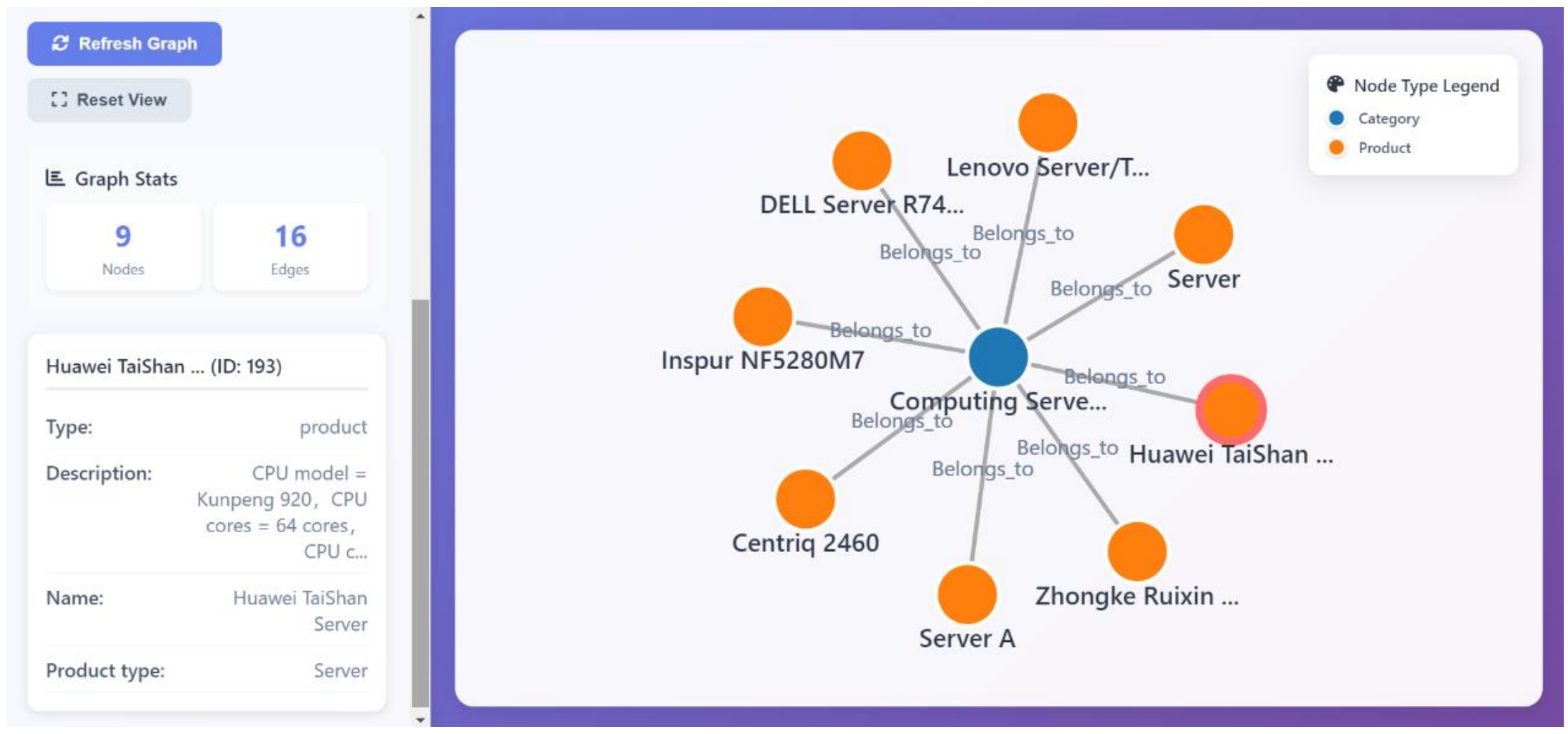

Figure 4. Example of KG Node Attribute Information Display.

### 4.3 Expand and Hide Nodes

#### 4.3.1 Expand Node

This function allows users to explore the KG step by step, on demand, avoiding visual clutter caused by loading too much information at once. As shown in Figure 5, after selecting the "Huawei TaiShan Server" product node, the user calls up the menu with a right-click and clicks "Expand Node". The system then dynamically loads and displays all first-degree neighbor nodes directly connected to this product, namely its brand "Huawei", model "Huawei TaiShan", and price "23500 yuan". This process is accompanied by smooth animation effects, providing users with clear context awareness.

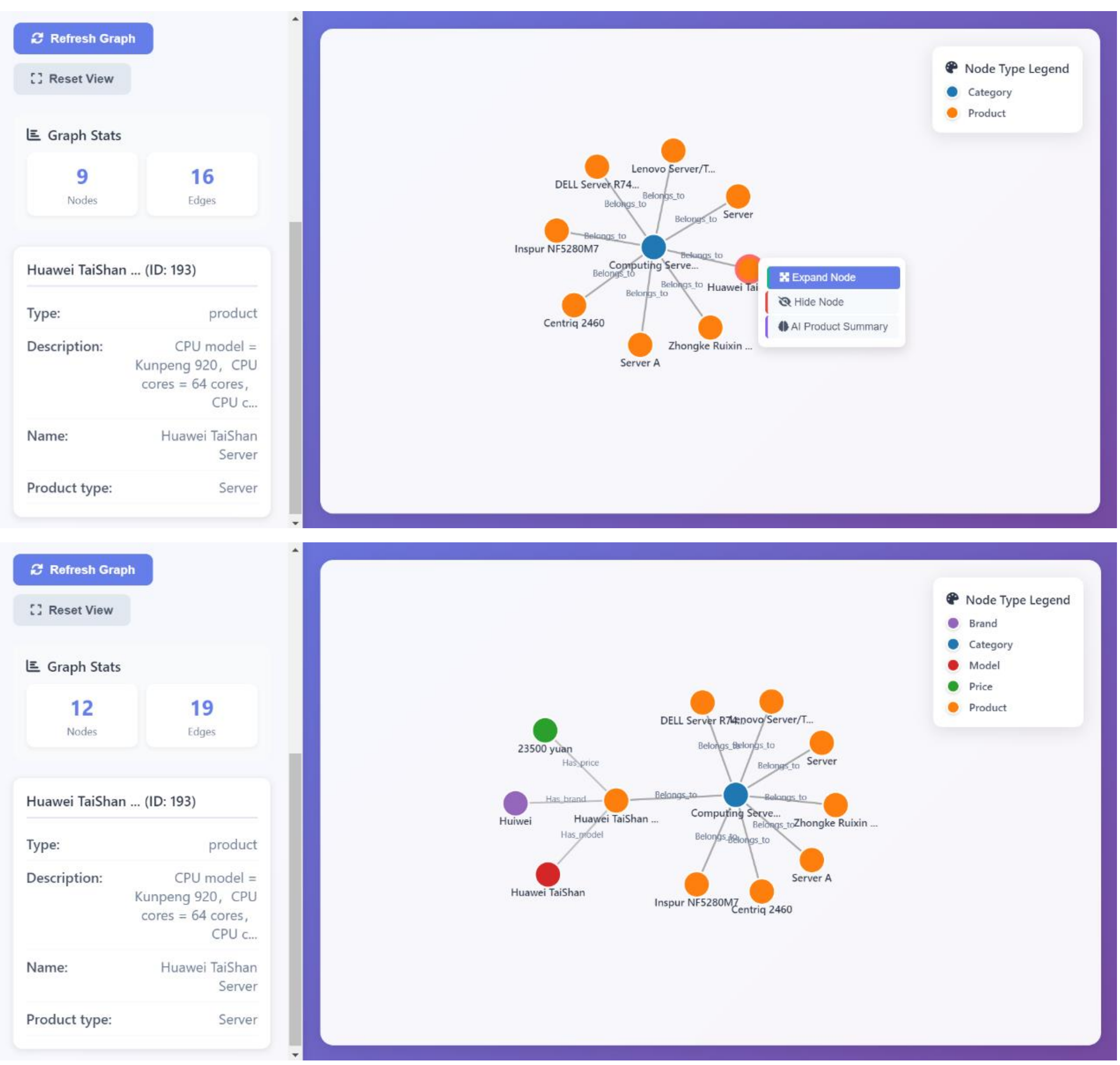

Figure 5. Example of KG Expand Node Interactive Function.

### 4.3.2 Hide Node

Corresponding to the expand node function, the hide node function allows users to actively simplify the view, remove information currently not of concern, and thus focus on the most important parts. As shown in Figure 6, the user may temporarily not care about other products under the "Huawei" brand, so they right-click the "Huawei" brand node and select "Hide Node". This node and all its connected edges are immediately removed from the view, making the graph structure more concise.

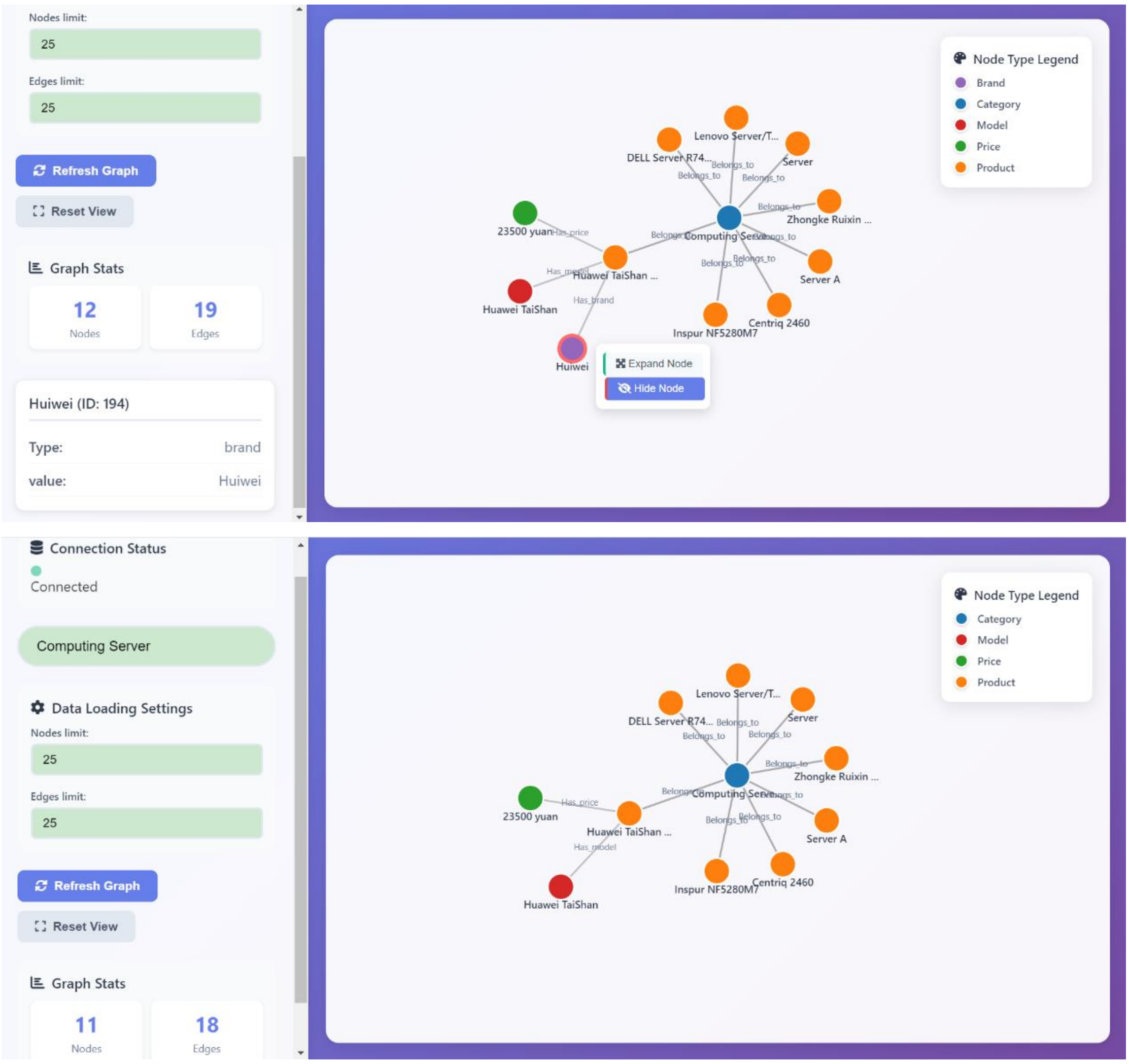

Figure 6. Example of KG Hide Node Interactive Function.

## 4.4 Agent Product Analysis

This is the key function of the method in this paper, seamlessly combining structured KGs with generative AI analysis. As shown in Figure 7, when the user is interested in gaining a deeper understanding of the "Huawei TaiShan Server", they can right-click the node and select "AI Product Summary". The system immediately extracts the key information of the product (name, brand, model), calls the integrated third-party LLM agent API in the background, and presents the detailed analysis report generated by the AI to the user in a pop-up window within seconds. The report content is comprehensive and professional, covering multiple dimensions such as product information overview, technical specifications, and application scenarios, greatly expanding the depth of the user's data analysis. This function successfully upgrades the KG from a "relationship displayer" to a "trigger for deep insights".

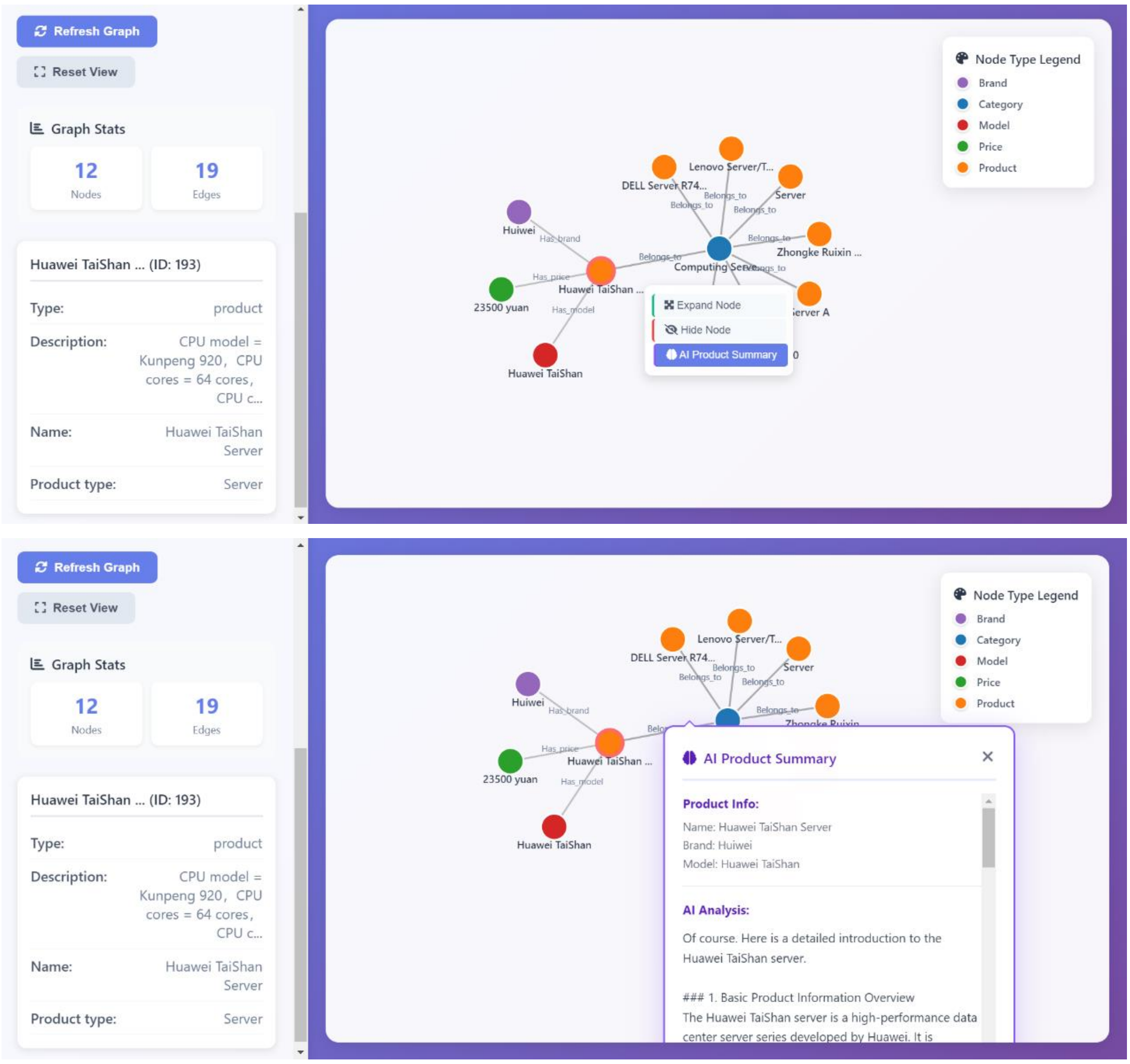

Figure 7. Example of KG AI Product Summary Interactive Function.

## 5. Discussion

### 5.1 Method Advantages and Innovativeness

The experimental results demonstrate the effectiveness and practicality of our proposed method. This method successfully creates a closed loop: the agent processes data to construct the KG, the KG is presented to the user through the visual interface, and the user's interactive behavior (especially the AI analysis request) in turn triggers real-time interaction with another AI agent (LLM). Specifically, the method proposed in this paper shows significant advantages and innovativeness in the following aspects:

(1) **Bidirectional Dynamic Interaction Mechanism:** In traditional methods, KGs are usually used as static knowledge bases retrieved or enhanced by LLMs. This paper breaks through this unidirectional pattern, achieving bidirectional dynamic interactions between LLM agents and KGs: the agent not only consumes knowledge but also actively constructs and extends the graph, forming a "construction-analysis-evolution" closed loop.

(2) **Support for Multi-dimensional Exploratory Analysis:** Existing methods mostly focus on single tasks (such as QA), while the framework in this paper supports users in conducting exploratory analysis of the product ecosystem from multiple perspectives and levels, achieving seamless connection from macro relationships to micro insights.

(3) **Deep Integration of Human-Machine Collaboration:** By seamlessly integrating interactive visualization with LLM agent analysis, users can directly trigger agents for deep analysis on the graph interface, achieving efficient human-machine collaboration of "what you see is what you analyze".

(4) **Fact Anchoring and Semantic Enhancement:** The KG provides structured factual basis for the LLM, effectively reducing the risk of model hallucination. Meanwhile, the LLM injects semantic understanding and reasoning capabilities into the graph, enhancing the depth and breadth of analysis.

### 5.2 Limitations and Future Work

Although the method in this paper shows potential in many aspects, the following limitations still exist and deserve further exploration in future work:

(1) **Knowledge Graph Construction and Update:** Currently, the acquisition of structured data and the construction of the KG are completed in two separate stages. How to achieve automated, real-time incremental updating of the KG is a challenge. Future work could research the use of more advanced LLM agents to continuously monitor data sources, automatically identify changes, and update the graph, forming a self-adaptive knowledge system.

(2) **Reliability of LLMs:** Although accuracy can be improved through prompt engineering and KG context injection, the LLM still carries the risk of "hallucination" [38]. Future work could consider introducing multi-model fusion mechanisms or building domain-specific lightweight models to enhance robustness.

(3) **Richness of Interaction Modes:** The current interactive functions are still mainly node expansion, hiding, and AI introduction. More analytical perspectives (such as temporal evolution, community discovery) and interaction methods (such as voice interaction, multimodal input) could be introduced in the future.

(4) **System Performance and Scalability:** When the scale of the KG expands to millions or even billions of levels, both the query efficiency of the backend and the rendering performance of the frontend will face challenges. Future research needs to focus on more efficient graph query algorithms, graph sampling or aggregation techniques, and higher-performance frontend rendering solutions based on technologies like WebGL.

(5) **Quantitative Evaluation:** The evaluation in this paper is mainly based on functional demonstrations and case analyses. A comprehensive evaluation framework needs to be established in the future, especially for such human-machine collaborative exploratory analysis systems. A series of standard analytical tasks can be designed to quantitatively evaluate the differences in the efficiency, depth, and novelty of insight discovery when users use this system compared to traditional data analysis tools [39][40].

## 6. Conclusion

Aiming at the problems of static interaction, single analytical dimension, and disjointed human-machine collaboration in current research on the integration of LLMs and KGs, this paper proposes a multi-dimensional data analysis method based on the interactions between LLM agents and KGs. By constructing an integrated framework containing four modules: data preparation, knowledge representation, visualization and interaction, and intelligent analysis, bidirectional dynamic interaction between the agent and the KG is achieved, enabling users to conduct deep exploration from macro structure to micro semantics on the visual platform.

Experimental results show that this method has significant effectiveness and practicality in product information analysis scenarios. It can not only automatically construct and extend the KG but also provide users with professional, in-depth product analysis reports through LLM agents, achieving an organic combination of knowledge representation and semantic reasoning.

This research provides a novel, flexible, and efficient solution for multi-dimensional data analysis, promoting the research on the integration of LLM agents and KGs towards a more dynamic, interactive, and collaborative direction. In the future, we will further optimize system performance, expand interaction modes, and explore its application potential in broader fields.